\documentclass{article}

\usepackage{microtype}
\usepackage{graphicx}
\usepackage{subfigure}
\usepackage{booktabs} 
\usepackage{enumitem}
\usepackage{hyperref}

\usepackage{tikz}
\usepackage{dsfont}
\usepackage{subcaption}

\usepackage[accepted]{icml2025}

\usepackage{amsmath}
\usepackage{amssymb}
\usepackage{mathtools}
\usepackage{amsthm}
\usepackage{tcolorbox}
\usepackage{dsfont}
\usepackage[capitalize,noabbrev]{cleveref}
\newcommand*\circled[1]{\tikz[baseline=(char.base)]{
            \node[shape=circle,draw,inner sep=1pt] (char) {#1};}}
\newcommand{\new}[1]{\textcolor{black}{#1}}

\theoremstyle{plain}
\newtheorem{theorem}{Theorem}[section]

\theoremstyle{definition}
\newtheorem{definition}[theorem]{Definition}

\theoremstyle{remark}

\definecolor{SP}{RGB}{30,160,156}

\usepackage[textsize=tiny]{todonotes}

\icmltitlerunning{Adaptive Learn-then-Test: Statistically Valid and Efficient Hyperparameter Selection}

\begin{document}

\twocolumn[
\icmltitle{Adaptive Learn-then-Test: Statistically Valid and Efficient Hyperparameter Selection}




\begin{icmlauthorlist}
\icmlauthor{Matteo Zecchin}{yyy}
\icmlauthor{Sangwoo Park}{yyy}
\icmlauthor{Osvaldo Simeone}{yyy}
\end{icmlauthorlist}

\icmlaffiliation{yyy}{Centre for Intelligent Information Processing Systems, Department of Engineering, King’s College London, London, United Kingdom}
\icmlcorrespondingauthor{Matteo Zecchin}{matteo.1.zecchin@kcl.ac.uk}
\icmlcorrespondingauthor{Sangwoo Park}{sangwoo.park@kcl.ac.uk}
\icmlcorrespondingauthor{Osvaldo Simeone}{osvaldo.simeone@kcl.ac.uk}
\icmlkeywords{Machine Learning, ICML}

\vskip 0.3in
]



\printAffiliationsAndNotice{} 

\begin{abstract}
We introduce adaptive learn-then-test (aLTT), an efficient hyperparameter selection procedure that provides finite-sample statistical guarantees on the population risk of AI models. Unlike the existing learn-then-test (LTT) technique, which relies on conventional p-value-based multiple hypothesis testing (MHT), aLTT implements sequential data-dependent MHT with early termination by leveraging e-processes. As a result, aLTT can reduce the number of testing rounds, making it particularly well-suited for scenarios in which testing is costly or presents safety risks. Apart from maintaining statistical validity, in applications such as online policy selection for offline reinforcement learning and \new{prompt engineering}, aLTT is shown to achieve the same performance as LTT while requiring only a fraction of the testing rounds. 
\end{abstract}

\section{Introduction}
\subsection{Context and Motivation}

\begin{figure*}
    \centering
    \includegraphics[width=0.95\linewidth]{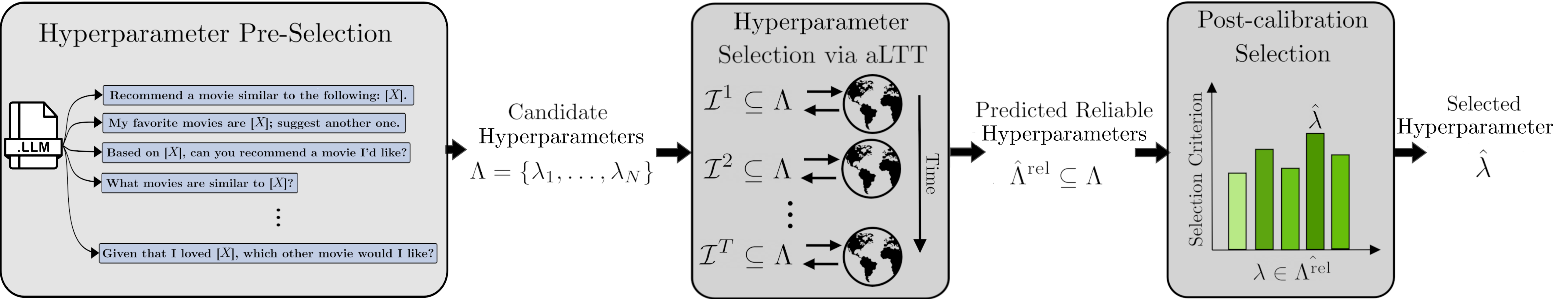}
    \vspace{-0.5em}
    \caption{\new{An example application of aLTT to reliable  prompt optimization   \cite{zhou2023large,quach2023conformal,schneider2024hyperband}. A set $\Lambda$ of candidate prompts  for a movie recommender  is generated using an LLM and/or prior experience. Prompts  serve as an example of a discrete set of hyperparameters to be optimized using aLTT. The goal is to select a subset  $\hat{\Lambda}^{\rm rel} \subseteq \Lambda$ of  prompts that guarantee a sufficiently high recommendation  accuracy.
    To this end, aLTT applies a sequence of data-dependent testing rounds with adaptive termination. Specifically, at  each testing round $t$, aLTT estimates the performance of a subset of hyperparameters $\mathcal{I}^t \subseteq \Lambda$ through held-out data or real-world testing. The subset to be tested is selected based on prior testing outcomes, and the process stops as soon as a sufficiently large reliable subset $\hat{\Lambda}^{\rm rel} $ is identified. An additional post-calibration selection step can be applied to choose a single hyperparameter $\hat{\lambda}$ from the selected subset $\hat{\Lambda}^{\rm rel}$ based on users preferences.} }
    \label{fig:pipeline}
    \vspace{-1em}
\end{figure*}

The safe and reliable deployment of AI applications, or apps for short, hinges on the possibility of certifying their performance \cite{seshia2022toward,tegmark2023provably}. Depending on the problem, this may require controlling the missed detection probability in medical imaging \cite{lu2022fair,mehrtash2020confidence}, ensuring safety measures for control policies \cite{lindemann2023safe,zecchin2024forking}, or verifying the correctness of the answers given by a large language model (LLM) \cite{quach2023conformal}.

In practice, before deployment, AI apps can be often \emph{calibrated} by selecting \emph{hyperparameters} based on data set aside for this purpose or based on rounds of real-world testing. \new{In the current scaling-centric era of large, Internet-scale, data sets \cite{xiao2024rethinking}, hyperparameter optimization is typically modeled as a bandit problem in which the goal is the minimization of the training loss over a set of \emph{discrete} configurations. The configurations may include prompt designs for the fine-tuning of large language models (LLMs) \cite{zhou2023large,quach2023conformal,schneider2024hyperband}, architectural choices related to model scale \cite{nasir2024llmatic} or to the timing of early decisions \cite{schuster2022confident}, or policy settings for offline  reinforcement learning \cite{paine2020hyperparameter,fujimoto2021minimalist}. }

\new{The learn-then-test (LTT) method introduced by \cite{ltt} has recently emerged as a formal framework to address hyperparameter selection over a discrete space.} LTT treats the calibration task of \emph{hyperparameter selection} as a \emph{multiple hypothesis testing} (MHT) problem. Accordingly, it associates each hyperparameter in a candidate set to the null hypothesis that the hyperparameter does not meet a reliability requirement on the population risk. Hypotheses are tested using p-values, and the probability of mistakenly detecting a hyperparameter as reliable is guaranteed via \emph{family-wise error rate} (FWER)-controlling statistical procedures \cite{keselman1977tukey}. This way, LTT ensures finite-sample, high-probability guarantees on the population risk of the selected hyperparameters. 

To prevent p-hacking \cite{head2015extent}, LTT’s guarantees apply only to \emph{non-adaptive} MHT procedures. However, related work on hyperparameter \emph{optimization} has shown the significant benefits that can be accrued by \emph{adaptive} exploration strategies that test hyperparameters sequentially in a data-driven manner \cite{swersky2014freeze,rakotoarison2024context}. This work aims at improving the data efficiency of LTT by leveraging recent advances in sequential MHT based on e-processes \cite{vovk2021values,waudby2024estimating,xu2021unified}.

\subsection{Related Work}
Finite-sample statistical guarantees on inferential outputs can be obtained via \textit{conformal prediction} methods (\citealp{shafer2008tutorial}; \citealp{angelopoulos2023conformal}) and, more generally, via conformal risk control and risk-controlling set-valued predictions \cite{angelopoulos2022conformal,bates2021distribution}. These methods calibrate classification or regression models by setting a threshold hyperparameter based on held-out data to control the size of a prediction set. Calibrated predictors can be leveraged in predict-then-optimize control tasks to offer reliability guarantees \cite{vovk2018conformal,lindemann2023safe,zecchin2024forking}.

Beyond prediction sets, the problem of calibrating an AI app via the selection of hyperparameters from a candidate pool has been addressed through the LTT framework \cite{ltt}. For example, \new{as illustrated  in  Figure \ref{fig:pipeline},  for the problem  of  prompt engineering \cite{zhou2023large}, the initial set of candidate hyperparameters 
 encompasses  instruction prompt templates generated by an LLM and/or via prior experience.} Leveraging p-value-based MHT via FWER-controlling procedures, LTT selects a subset of candidates that come with high-probability population risk guarantees.

FWER guarantees are often too conservative, potentially resulting in empty calibration sets. The \textit{false discovery rate }(FDR) is an alternative and less strict criterion that is often preferred for MHT in fields such as genetics \cite{van2003false}, neuroimaging \cite{genovese2002thresholding}, online advertising \cite{berman2022false}, and finance \cite{harvey2020false}.

\textit{E-value}s have gained popularity in MHT due to their advantages over p-values \cite{vovk2021values,shafer2019game,ramdas2020admissible}. Similarly to p-values, e-values measure the statistical plausibility of the null hypothesis. Specifically, an e-value can be thought of as a special type of p-value that has additional robustness properties. Specifically, the key property of interest in this paper is that, unlike p-values, e-values can be readily combined to obtain \textit{e-processes}, making it possible to devise \textit{sequential} testing strategies with \textit{anytime safety} \cite{wang2022false,ramdas2023game}. E-processes have been applied to problems such as sequential change detection \cite{shin2022detectors}, multiple bandit testing \cite{xu2021unified}, two-sample testing \cite{shekhar2023nonparametric}, and mean estimation of bounded random variables \cite{waudby2024estimating}.

\emph{Hyperparameter optimization} is a vast field focused on the optimization of the hyperparameters of training algorithms, such as learning rate, weight decay, and dropout rate  \cite{swersky2014freeze,pedregosa2016hyperparameter,maclaurin2015gradient,lindauer2022smac3}. Hyperparameter optimization typically operates in continuous domains, and it can serve as a preliminary step for the identification of candidate hyperparameters. While hyperparameter optimization does not provide statistical guarantees on the population risk, the goal of hyperparameter selection methods such as LTT is to formally test a subset of candidate hyperparameters for statistical validity.

\subsection{Main Contributions}
The main contributions of this paper are as follows. 

    $\bullet$ We introduce adaptive LTT (aLTT), a data-efficient hyperparameter selection method that provides finite-sample guarantees on the population risk of AI apps. The main technical underpinning of aLTT is \textit{e-process-based MHT}, which supports statistical validity while enabling data-dependent sequential testing \cite{xu2021unified}. Unlike LTT, as illustrated in Figure \ref{fig:pipeline}, aLTT adaptively tests subsets of hyperparameters that are chosen based on the evidence accumulated in the previous rounds, allowing also for the early termination of the calibration process. aLTT guarantees rigorous control over FWER and FDR, while significantly reducing the number of testing rounds.
    
    $\bullet$ We study two practical scenarios requiring hyperparameter selection, namely online policy selection for offline reinforcement learning \cite{fujimoto2021minimalist} and \new{automated prompt engineering} \cite{zhou2023large}. In both cases, aLTT is shown to deliver reliable and effective hyperparameters using only a small fraction of the testing rounds required by LTT. 

\section{Problem Definition}
\subsection{Setting}
Let $\mathcal{M}_\lambda$ be an AI app whose operation is determined by a vector of hyperparameters $\lambda$.  The performance of a hyperparameter vector $\lambda$ when tested at input data $Z$ is measured by a risk function $R(\lambda,Z)\in[0,1]$. Accordingly, the \textit{population risk} with respect to an unknown data distribution $P_{Z}$ is defined as
\begin{align}
	R(\lambda)=\mathbb{E}_{P_{Z}}[R(\lambda,Z)].
	\label{eq:pop_risk}
\end{align} 

As illustrated in Figure \ref{fig:pipeline}, \new {for a given discrete subset $\Lambda=\{\lambda_1,\dots,\lambda_N\}$ of hyperparameters and a user-specified reliability level $\alpha\in [0,1]$,} we aim at determining the subset of hyperparameters in set $\Lambda$ that conforms with the required reliability level $\alpha$, i.e.,
\begin{align}
	\Lambda^{\rm rel}=\{\lambda\in \Lambda: R(\lambda)\leq \alpha \}.
	\label{eq:rel_set}
\end{align}
The complementary set, comprising unreliable hyperparameters, is accordingly defined as
\begin{align}
	\Lambda^{\rm unrel}=\Lambda\setminus\Lambda^{\rm rel}=\{\lambda\in \Lambda: R(\lambda)>\alpha \}.
	\label{eq:unrel_set}
\end{align}
Since identifying the entire set $\Lambda^{\rm rel}$ in (\ref{eq:rel_set}) is impossible owing to the lack of knowledge about the data distribution $P_Z$, the goal is producing a subset of hyperparameters $\hat{\Lambda}^{\rm rel}\subseteq \Lambda$  that contains as many reliable hyperparameters from subset $\Lambda^{\rm rel}$ as possible while controlling the number of unreliable hyperparameters from subset $\Lambda^{\rm unrel}$ mistakenly included in subset $\hat{\Lambda}^{\rm rel}$.

\subsection{Performance Criteria}
\label{sec:perf_crit}
\begin{definition}[$(\alpha,\delta)$-FWER-controlling set]
	For a given reliability level $\alpha\in[0,1]$ and an error level $\delta\in [0,1]$, a hyperparameter subset $\hat{\Lambda}^{\rm rel}\subseteq\Lambda$ is $(\alpha,\delta)$-FWER-controlling set if it satisfies the requirement
	\begin{align}
		{\rm FWER}(\hat{\Lambda}^{\rm rel}):=\Pr\left[|\Lambda^{\rm unrel}\cap\hat{\Lambda}^{\rm rel}|\geq1\right]\leq \delta. 
		\label{eq:FWER_guarantee}
	\end{align}
	where the probability is evaluated with respect to the distribution of the subset $\hat{\Lambda}^{\rm rel}$.
\end{definition}

The FWER guarantee (\ref{eq:FWER_guarantee}) imposes that the probability that the calibration set $\hat{\Lambda}^{\rm rel}$ contains an unreliable hyperparameter is bounded by $\delta$.

\begin{definition}[$(\alpha,\delta)$-FDR-controlling set]
	For a given reliability level $\alpha\in[0,1]$ and an error level $\delta\in [0,1]$, a hyperparameter subset $\hat{\Lambda}^{\rm rel}\subseteq\Lambda$ is $(\alpha,\delta)$-FDR-controlling set if it satisfies the inequality
	\begin{align}
		{\rm FDR}(\hat{\Lambda}^{\rm rel}):=\mathbb{E}\left[\frac{|\Lambda^{\rm unrel}\cap\hat{\Lambda}^{\rm rel}|}{|\hat{\Lambda}^{\rm rel}|}\Bigg||\hat{\Lambda}^{\rm rel}|\geq1\right]\leq \delta, 
		\label{eq:FDR_guarantee}
	\end{align}
	with the average evaluated with respect to the distribution of the subset $\hat{\Lambda}^{\rm rel}$.
\end{definition}

Accordingly, a testing procedure that outputs $(\alpha,\delta)$-FDR-controlling sets guarantees that the expected fraction of unreliable hyperparameters in the predicted set $\hat{\Lambda}^{\rm rel}$ is bounded by $\delta$. Thus, ensuring the $(\alpha,\delta)$-FWER condition automatically also guarantees the $(\alpha,\delta)$-FDR  requirement.

Since any FWER or FDR level can be trivially satisfied by a procedure that returns the empty set $\hat{\Lambda}^{\rm rel} = \emptyset$, it is important to gauge the informativeness of the testing procedure via the \emph{true positive rate} (TPR), which corresponds to the expected fraction of reliable models in the predicted set $\hat{\Lambda}^{\rm rel}$, i.e.,
\begin{align}
	{\rm TPR}(\hat{\Lambda}^{\rm rel})=\mathbb{E}\left[\frac{|\Lambda^{\rm rel}\cap \hat{\Lambda}^{\rm rel}|}{\left|\Lambda^{\rm rel}\right|}\right].
\end{align}

\subsection{Sequential and Adaptive Hyperparameter Selection}
\label{sec:iterative_procedure}
To produce the estimated subset of reliable hyperparameters, $\hat{\Lambda}^{\rm rel}$, we adopt a general sequential testing procedure that, at each round $t\geq1$, operates as follows.
\begin{tcolorbox}[right=0.8em,left=0.8em,top=0.2em,bottom=0.2em]
\begin{enumerate}[leftmargin=*,labelindent=0pt,wide, labelwidth=0pt,label=\protect\circled{\arabic*}]
	\item \label{step_1} \emph{Hyperparameter subset selection:} A subset of hyperparameters $\mathcal{I}^t \subseteq \Lambda$ is selected for testing.
	\item \emph{Testing:} Empirical risk estimates $\mathcal{R}^t = \{R(\lambda_i, Z_i^t)\}_{\lambda_i \in \mathcal{I}^t}$ are obtained, one for each candidate hyperparameter $\lambda_i$ in the selected subset $\mathcal{I}^t$, using held-out data or real-world testing. The random variable $Z^{t}_i \sim P_Z$ describes the data used to test hyperparameter $\lambda_i$ at round $t$. The random variables $\{Z_i^t\}_{i\in \mathcal{I}^t}$ can be arbitrarily dependent, and thus one may reuse the same data to test all hyperparameters $\lambda\in\mathcal{I}^t$. 
	\item \emph{Evidence update:} Evidence accumulated up to time $t$, including both the observed risks and the subset of queried models, is updated as $\mathcal{D}^t = \mathcal{D}^{t-1}\cup \{(\mathcal{I}^t, \mathcal{R}^t)\}$. 
\end{enumerate}
\end{tcolorbox}

The testing procedure outlined above is fully specified by the tuple $\Pi=(\{\mathcal{Q}^t\}_{t\geq1}, \mathcal{A}, T)$, encompassing a family of acquisition policies $\{\mathcal{Q}^t\}_{t\geq1}$, a decision rule $\mathcal{A}$, and a calibration horizon $T$, which are defined as follows.

$\bullet$ \emph{Acquisition policy:} At each round $t$, the acquisition policy $\mathcal{Q}^t$ determines the hyperparameters $\mathcal{I}^t$ to be tested at step \ref{step_1}. If the policy $\mathcal{Q}^t$ uses the evidence $\mathcal{D}^{t-1}$ to select the hyperparameter set $\mathcal{I}^t$, it is said to be \emph{adaptive}; otherwise, it is \emph{non-adaptive}. Both \emph{adaptive} and \emph{non-adaptive} acquisition policies can incorporate prior knowledge, which we denote as $\mathcal{D}^{0}$.

$\bullet$ \emph{Decision rule:}  The decision rule $\mathcal{A}$ uses the evidence $\mathcal{D}^{T}$ available at the end of the last calibration round $T$ to produce the estimated set $\hat{\Lambda}^{\rm rel}$ of reliable hyperparameters.

$\bullet$  \emph{Number of calibration rounds:}  The number of calibration rounds $T$, is said to be \emph{adaptive}, if the stopping condition $T=t$ is determined by the evidence $\mathcal{D}^{t-1}$. Otherwise, when it is predetermined based solely on prior knowledge $\mathcal{D}^{0}$, the calibration horizon $T$ is said to be \emph{non-adaptive}.

\section{(Non-Adaptive) Learn-then-Test}
\label{sec:batchltt}

In this section, we review LTT, a \emph{non-adaptive} hyperparameter selection procedure devised to meet the $(\alpha,\delta)$-FWER guarantee \cite{ltt}. LTT associates to each hyperparameter $\lambda_i\in\Lambda$ the null hypothesis
\begin{align}
	\mathcal{H}_i: R(\lambda_i)>\alpha
	\label{eq:hypothesis}
\end{align} 
that the population risk $R(\lambda_i)$ in (\ref{eq:pop_risk}) violates the target reliability level $\alpha$. For each null hypothesis $\mathcal{H}_i$ a p-value $P_i$ is a non-negative random variable that satisfies the inequality
\begin{align}
	\Pr[P_i\leq x|\mathcal{H}_i]\leq x
	\label{eq:p_value}
\end{align}
for every $x\in[0,1]$. By the definition (\ref{eq:p_value}) a p-value $P_i$ provides evidence for the validity of hypothesis $\mathcal{H}_i$. This is in the sense that a small value of $P_i$ is unlikely to occur if $\mathcal{H}_i$ is true.

Given a pre-specified calibration horizon $T$ and a non-adaptive acquisition policy $\{Q^t\}_{t\geq1}$, LTT queries at each round $t\geq 1$ the subset of hyperparameters $Q^t(\mathcal{D}^0)=\mathcal{I}^t$, obtaining the corresponding risk estimates $\mathcal{R}^t=\{R(\lambda_i, Z_i^t)\}_{\lambda_i\in \mathcal{I}^t}$.

LTT uses the accumulated evidence at the end of the testing process, $\mathcal{D}^T$, to compute a valid p-value for the null hypothesis (\ref{eq:hypothesis}) using, for instance, the Hoeffding-Bentkus concentration inequality introduced in \cite{bates2021distribution}. Based on the collection of p-values $\mathcal{P} = \{P_i\}_{i=1}^N$,  LTT selects a subset of hyperparameters $\hat{\Lambda}^{\rm LTT}$ using a FWER-controlling algorithm $\mathcal{A}^{\rm FWER}(\mathcal{P})$. A variant of LTT that is FDR-controlling can be readily obtained by using an FDR-controlling selection rule $\mathcal{A}^{\rm FDR}(\mathcal{P})$. Examples of FWER and FDR controlling procedures are provided in the Supplementary Material.
\begin{algorithm*}
\caption{Adaptive Learn-Then-Test (aLTT)}\label{alg:cap}
\begin{algorithmic}
\STATE{\bfseries Require:}  Candidate hyperparameters $\Lambda$, prior knowledge $\mathcal{D}^0$, reliability level $\alpha$, error tolerance level $\delta$, acquisition policy $\{\mathcal{Q}^t\}_{t\geq 1}$, FWER/FDR-controlling selection rule $\mathcal{A}^{\rm FWER}$/$\mathcal{A}^{\rm FDR}$, betting strategy $\{\mu^t_i\}_{t\geq 1,i=1,\dots,N}$, maximum number of iterations $t_{\rm max}$ and minimal hyperparameter set cardinality $d$
\STATE{\bfseries Ensure:}  $(\alpha,\delta)$-FWER/FDR-controlling hyperparameter set $\hat{\Lambda}^{{\rm aLTT},T}$
\STATE $t\leftarrow 1$
\WHILE{$t\leq t_{\rm max} \land |\hat{\Lambda}^{{\rm aLTT},t}|\leq d$}
\STATE Select hyperparameters $\mathcal{I}^t=\mathcal{Q}^t(\mathcal{E}^{t-1})$ and receive risk estimates $\mathcal{R}^t= \{R(\lambda_i, Z_i^t)\}_{\lambda_i \in \mathcal{I}^t}$
\STATE Update evidence $\mathcal{D}^t = \mathcal{D}^{t-1}\cup \{(\mathcal{I}^t, \mathcal{R}^t)\}$ and e-processes $\mathcal{E}^{t}$ as in (\ref{eq:eproc_update})
\IF{FWER-control}
    \STATE Compute p-values $\mathcal{P}^t$ as in (\ref{eq:villespvalues})
    \STATE $\hat{\Lambda}^{{\rm aLTT},t}\leftarrow\mathcal{A}^{\rm FWER}(\mathcal{P}^t)$
\ELSIF{FDR-control}
    \STATE $\hat{\Lambda}^{{\rm aLTT},t}\leftarrow\mathcal{A}^{\rm FDR}(\mathcal{E}^t)$
\ENDIF
\STATE $t\leftarrow t+1$
\ENDWHILE
\STATE{\bfseries Return: }$\hat{\Lambda}^{{\rm aLTT},T}$
\end{algorithmic}
\label{alg:altt}
\end{algorithm*}

\section{Adaptive Learn-Then-Test}
\label{sec:seqltt}

In this section, we introduce \emph{adaptive LTT} (aLTT), a hyperparameter selection scheme that supports \emph{adaptive} acquisition policies and an \emph{adaptive} number of calibration rounds. The algorithmic description of aLTT is given in Algorithm \ref{alg:altt}.

\subsection{Hypothesis Testing via E-Processes}
\label{sec:seq_mht}
The proposed aLTT scheme applies MHT based on e-values and e-processes \cite{shafer2021testing,ramdas2023game}. For each null hypothesis $\mathcal{H}_i$ in (\ref{eq:hypothesis}), an \textit{e-value} $E_i$ is a non-negative random variable with an expectation no larger than 1 when $\mathcal{H}_i$ is true, i.e.,
\begin{align}
	\mathbb{E}[E_i|\mathcal{H}_i]\leq 1.
	\label{eq:evalue}
\end{align}
By Markov's inequality, an e-value $E_i$ can be turned into a p-value $P_i$ as $P_i= 1/E_i$, since the inequality (\ref{eq:p_value}) is satisfied as 
\begin{align}
	\Pr\left[\frac{1}{E_i}\leq x\bigg|\mathcal{H}_i\right]\leq \mathbb{E}[E_i|\mathcal{H}_i]x\leq x, {\color{black} \text{ }\forall x \in [0,1].}
\end{align}

For each null hypothesis $\mathcal{H}_i$, given an observation $Z$ and a fixed $\mu\in (0,1/(1-\alpha))$, a valid e-value is given by \cite{waudby2024estimating}
\begin{align}
	\label{eq:value_hi}
	E_i=(1+\mu(\alpha-R(\lambda_i,Z))).
\end{align}
The e-value (\ref{eq:value_hi}) has the interpretation of \textit{wealth growth} in a \textit{betting} setting. Accordingly, one can think of parameter $\mu$ in (\ref{eq:value_hi}) as the amount of the current wealth that the gambler bets on the hypothesis $\mathcal{H}_i$ being false, i.e., on the validity of the assumption $R(\lambda_i)\leq \alpha$ that the hyperparameter $\lambda_i$ is reliable. In fact, if $\mu>0$, when $R(\lambda_i,Z)\leq \alpha$, the gambler's wealth in (\ref{eq:value_hi}) increases; while, when $R(\lambda_i,Z)> \alpha$, the quantity (\ref{eq:evalue}) the gambler's wealth (\ref{eq:value_hi}) decreases. 

An e-process for hypothesis $\mathcal{H}_i$ is a sequence of random variables $\{E_i^t\}_{t\geq1}$ such that, for any stopping time $T$, which may depend on all previously collected evidence, the random variable $E_i^T$ is a valid e-value.
Using the e-value (\ref{eq:value_hi}), considering the general iterative testing framework in Section \ref{sec:iterative_procedure}, an e-process for the null hypothesis $\mathcal{H}_i$ in (\ref{eq:hypothesis}) can be obtained as the product
\begin{align}
	E_i^t=\prod_{\tau\leq t: \lambda_i\in \mathcal{I}^\tau}(1+\mu_i^\tau(\alpha-R(\lambda_i,Z^\tau_i)))
	\label{eq:cap_wealth},
  \vspace{-1em}
\end{align}
where the betting strategy $\mu_i^t\in (0,1/(1-\alpha))$ can be optimized as a function of the past risk estimates $\{R(\lambda_i,Z^{\tau}_i)\}_{\tau<t}$ and $E^0_i=1$. Based on the discussion above, the e-process (\ref{eq:cap_wealth}) represents the wealth accumulated up to time $t$ by a gambler making sequential bets $\{\mu^\tau_i\}_{\tau<t}$ \cite{shafer2019game,waudby2024estimating}. As such, the gambler’s wealth up to time $t$, $E^t_i$, provides evidence against the null hypothesis that the hyperparameter $\lambda_i$ is unreliable. 

In a similar way, an \emph{anytime-valid p-value} for hypothesis $\mathcal{H}_i$ is a sequence of random variables $\{P_i^t\}_{t\geq1}$ such that, for any stopping time $T$, the random variable $P_i^T$ is a valid p-value. Given an e-process $\{E_i^t\}_{t\geq1}$ for hypothesis $\mathcal{H}_i$, the sequence
\begin{align}
	P^t_i =  \frac{1}{\max_{\tau \leq t} E^\tau_i}
	\label{eq:villespvalues}
\end{align} 
is an anytime-valid p-value for the hypothesis $\mathcal{H}_i$ \cite{ramdas2023game}.

\subsection{Adaptive Acquisition Policy}
 aLTT applies an \emph{adaptive} acquisition policy $\{\mathcal{Q}^t\}_{t\geq 1}$ and  an \emph{adaptive} calibration horizon $T$. Specifically, at each calibration round $t \geq 1$, aLTT's acquisition policy $\mathcal{Q}^t$ uses the  e-processes $\mathcal{E}^{t-1}=\{E_i^{t-1}\}_{i=1}^N$ in (\ref{eq:cap_wealth}) to choose which subset of hyperparameters, $\mathcal{I}^t$, to test next. Examples of acquisition functions $\mathcal{I}^t=\mathcal{Q}^t(\mathcal{E}^{t-1})$ will be provided  in the next section.

For the selected hyperparameters in set $\mathcal{I}^t$, aLTT obtains the risk estimates $\mathcal{R}^t=\{R(\lambda_i, Z_i^t)\}_{\lambda_i\in \mathcal{I}^t}$  and updates the associated e-processes using the recursive formula (\ref{eq:cap_wealth}), i.e.,
\begin{align}
	\label{eq:eproc_update}
	E^{t}_i=\begin{cases}
		(1+\mu_i^t(\alpha-R(\lambda_i,Z^t_i)))E^{t-1}_i, \quad &\text{if $\lambda_i\in \mathcal{I}^t$}\\
		E^{t-1}_i, \quad& \text{otherwise}.
	\end{cases}
  \vspace{-1em}
\end{align}
With this information, a prediction set $\hat{\Lambda}^{{\rm aLTT},t}$ is evaluated by employing either an FWER-controlling method $\mathcal{A}^{\rm FWER}(\mathcal{P}^t)$ based on the p-values $\mathcal{P}^t = \{P^t_i\}_{i=1}^N$ in (\ref{eq:villespvalues}); or an FDR-controlling procedure $\mathcal{A}^{\rm FDR}(\mathcal{E}^t)$, such as the e-Benjamini-Hochberg (eBH) method, reviewed in the Supplementary Material \cite{wang2022false}, using directly the e-values (\ref{eq:eproc_update}).

aLTT terminates the calibration procedure whenever there are at least $d$ hyperparameters in set $\hat{\Lambda}^{{\rm aLTT},t}$, i.e., $|\hat{\Lambda}^{{\rm aLTT},t}|\geq d$, or a maximum number of iterations $t_{\rm max}$ have been reached. This allows aLTT to stop the data acquisition phase early when a sufficiently large number of reliable hyperparameters have been identified.

\subsection{Hyperparameter Subset Selection}
At the end of the calibration process, aLTT uses the current e-processes $\mathcal{E}^{T}$ to generate the final prediction set $\hat{\Lambda}^{{\rm aLTT},T}$. By the anytime validity properties explained in Section \ref{sec:seq_mht}, if an FWER-controlling method $\mathcal{A}^{\rm FWER}(\mathcal{P}^T)$ is used, the resulting set  $\hat{\Lambda}^{{\rm aLTT},T}$ is  $(\alpha,\delta)$-FWER-controlling; while if an FDR-controlling method is used, the resulting set $\hat{\Lambda}^{{\rm aLTT},T}=\mathcal{A}^{\rm FDR}(\mathcal{E}^T)$ is $(\alpha,\delta)$-FDR-controlling.

\section{Applications}  
\label{sec:exp}   
\subsection{Online Policy Selection for  Offline Reinforcement Learning}
Offline reinforcement learning enables the training of control policies based on a fixed data set collected by using a possibly unknown behavior policy, without any online interaction with the environment \cite{levine2020offline}. However, the estimate of the performance of the trained policies obtained from offline data can differ substantially from the actual performance in the real world. This makes it practically essential to validate the policies’ performance via online interaction with the environment \cite{paine2020hyperparameter,liu2023offline}.

To reduce the cost and potential harm of online validation procedures, the number of online interactions of the pre-trained candidate policies with the real world must be kept to a minimum \cite{garcia2015comprehensive}. To this end, in this subsection, we investigate the potential benefits of the proposed aLTT framework as a means to select a subset of candidate policies that enjoy performance guarantees with respect to the real-world environment. 
\begin{figure}
	\centering
	  \includegraphics[width=\linewidth]{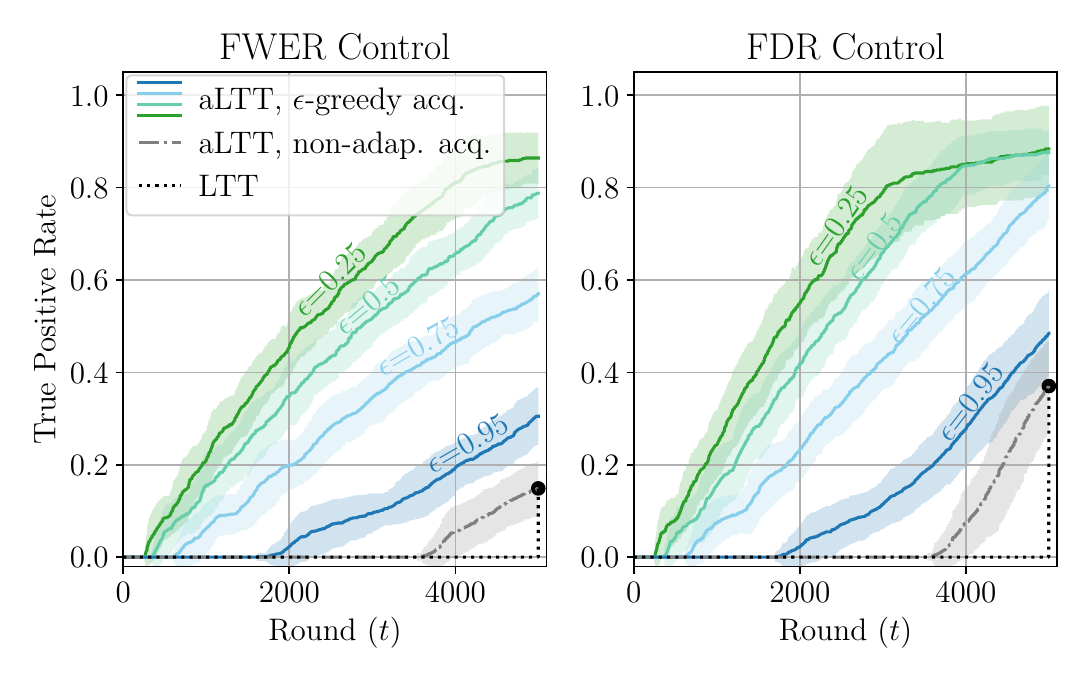}
        \vspace{-1em}
		\caption{True positive rate of LTT and aLTT with $\epsilon$-greedy acquisition policy for $\epsilon\in\{0.25,0.5,0.75,0.95\}$ and non-adaptive acquisition. On the left panel, the prediction sets satisfy FWER control while on the right FDR control. In both cases, the tolerance level is $\delta=0.1$. }
	\label{fig:rl_res}
  \vspace{-1em}
\end{figure}
\subsubsection{Problem Definition}
We assume a standard Markov decision process (MDP) $\mathcal{E}=\{\mathcal{S},\mathcal{A},P_{s'|s,a},P_{r|s,a}\}$ specified by a state space $\mathcal{S}$; an action space $\mathcal{A}$; a transition kernel $P_{s'|s,a}$, defining the conditional distribution of the next state $s'\in \mathcal{S}$ given the current state $s\in\mathcal{S}$ and action $a\in \mathcal{A}$; and a conditional reward distribution $P_{r|s,a}$ given state $s\in\mathcal{S}$ and action $a\in \mathcal{A}$. We assume that the reward $r$ is bounded and normalized in the $[0,1]$ interval.

We are given a set of pre-trained control policies $\Pi=\{\pi_1,\dots,\pi_N\}$, mapping an observed state $s\in \mathcal{S}$ to the random action  $a\sim \pi_i(s)\in \mathcal{A}$. Each policy $\pi_i$ is identified by a hyperparameter $\lambda_i$. The goal is to select a subset $\hat{\Lambda}^{\rm rel}\subset \Lambda$ of hyperparameters that yield reliable policies by using a limited amount of interactions with the real world. 

Reliability is measured via the cumulative reward obtained by a policy $\lambda$ on the MDP $\mathcal{E}$, which is defined as
\begin{align}
	R(\lambda,Z)=\frac{1}{K}\sum^{K}_{k=1}r^k,
	\label{eq:average_reward}
\end{align}
where $K$ is the length of the episode, and the per-episode random variable $Z$ encompasses the initial state $s^1\sim P_{s^1}$ along with the sequence $a^1,r^1,s^2,a^2,r^2,\dots,s^K,a^K,r^K$ with actions $a^k\sim \pi_i(s^k)$, rewards $r^k\sim P_{r^k|s^k,a^k}$ and MDP transitions $s^{k+1}\sim P_{s^{k+1}|s^k,a^k}$. 
For a user-specified reliability level $\alpha\in(0,1)$, the subset of reliable policies $\Lambda^{\rm rel}\subseteq \Lambda$ includes all policies in $\Lambda$ with average cumulative reward larger than $\alpha$, $\Lambda^{\rm rel}=\{\lambda\in\Lambda: R(\lambda)=\mathbb{E}_{P_Z}[R(\lambda,Z)]> \alpha\}$, while the complementary set $\Lambda^{\rm unrel}=\Lambda\setminus \Lambda^{\rm rel}$ includes the policies that do not satisfy the given reliability requirement.

Control policies are tested sequentially by following the procedure described in Section \ref{sec:iterative_procedure}, such that at each calibration round $t$ a policy $\lambda^t_i\in \Lambda$ is tested using online interactions with the MDP $\mathcal{E}$ to obtain the episodic reward value $R^t_i=R(\lambda^t_i,Z^t_i)$ in (\ref{eq:average_reward}). For an error threshold $\delta\in[0,1]$, the goal of reliable online policy selection is to return a prediction set $\hat{\Lambda}^{\rm rel}\subseteq\Lambda$ that is either $(\alpha,\delta)$-FDR controlling or $(\alpha,\delta)$-FWER controlling with a TPR that is as large as possible.

\begin{figure}
	\centering
		\includegraphics[width=0.8\linewidth]{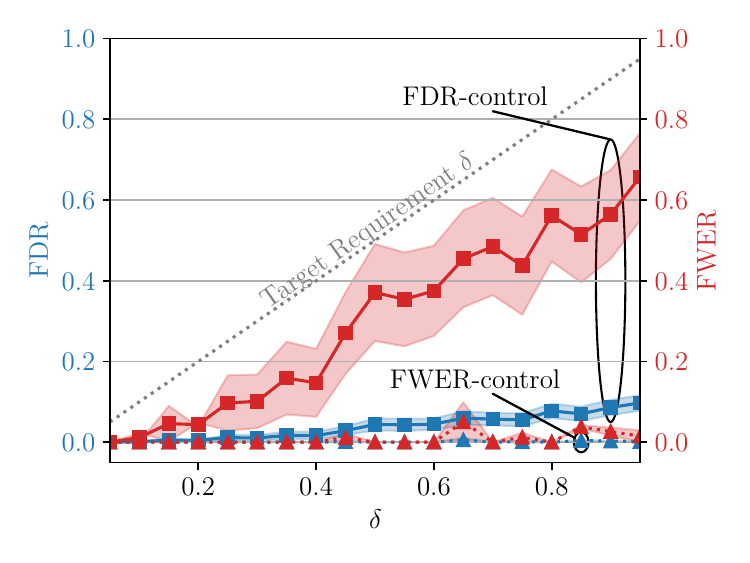}  
        \vspace{-1em}
		\caption{Comparison of the FWER and FDR levels obtained by aLTT under FDR-control (solid lines) and FWER-control (dashed lines) for different maximum tolerated error (FWER or FDR) levels $\delta$.}
  \label{fig:rl_delta}
   \vspace{-1em}
\end{figure}

\subsubsection{Results}
In our experiments, we consider the Half Cheetah control problem from the OpenAI Gym MuJoCo tasks \cite{todorov2012mujoco} and use control policies obtained via the offline reinforcement learning algorithm TD3+BC \cite{fujimoto2021minimalist}. The TD3+BC algorithm leverages an offline data set $\mathcal{D}$ to optimize policies by maximizing the standard deterministic policy gradient objective \cite{silver2014deterministic}, combined with a behavioral cloning regularization term, whose strength is controlled by a hyperparameter $\lambda$ in [Eq. 5]\cite{silver2014deterministic}. We produce $N=20$ different control policies by setting the hyperparameter $\lambda$ in the TD3+BC training objective on an evenly spaced grid in the interval $[0.25, 5]$. Unless stated otherwise, we consider a target reliability $\alpha=0.57$ and a target FDR requirement $\delta=0.1$. 

We evaluate aLTT with an $\epsilon$-greedy acquisition policy $Q^{t}$ that, at every calibration round $t$, with probability $1-\epsilon$, selects the hyperparameter $\lambda^t_i$ not included in $\hat{\Lambda}^{\rm aLTT,t}$ that is associated with the largest e-process value; otherwise, it picks uniformly at random a hyperparameter not in $\hat{\Lambda}^{\rm aLTT,t}$. For reference, we also consider aLTT with a non-adaptive acquisition policy that, at each round $t$, picks uniformly at random the hyperparameter to be tested regardless of the prediction outcome $\hat{\Lambda}^{\rm aLTT,t}$ and the e-process values. As a benchmark, we implement LTT with a random uniform acquisition policy and p-values obtained from the e-processes as in (\ref{eq:villespvalues}). Recall that LTT produces a decision at the end of the calibration process, here at round $T=5000$.

Finally, the value of the parameter $\mu^{t}_i$ in aLTT is set by following the  \emph{approximate growth rate adaptive to the particular alternative} (aGRAPA) betting strategy in \cite{waudby2024estimating} with other adaptive and non-adaptive betting strategies evaluated in the Supplementary Material.

In Figure \ref{fig:rl_res}, we compare the TPR of LTT and aLTT as a function of the calibration round $t$. We target FWER control on the left and FDR control on the right. By construction, LTT returns uninformative hyperparameter sets up until the termination of the testing procedure. The performance of LTT is the same as aLTT with a non-adaptive acquisition policy, i.e. with $\epsilon=1$, at $t=T$.  aLTT with an $\epsilon$-greedy acquisition function can benefit from the accumulated evidence to adaptively determine the models to test next. As $\epsilon$ decreases, and thus the acquisition policy becomes increasingly driven by evidence, the TPR increases from 0.32 to 0.85 in the case of FWER control and from 0.4 to 0.85 in the case of FDR control. Finally, we note that the TPR of the schemes under FDR control is larger than that obtained under FWER control, reflecting the stricter reliability requirement of FWER control.

In Figure \ref{fig:rl_delta}, we report the FWER and FDR of aLTT with $\epsilon = 0.25$ as a function of the tolerated error level $\delta$. As the error level $\delta$ increases, the empirical FWER and FDR increase accordingly, remaining below the maximum target level $\delta$. However, since FWER control is a stricter requirement, the aLTT prediction set obtained under FWER control delivers lower FDR and FWER levels as compared to aLTT with FDR control.

{\color{black}

\subsection{Reliable Automated Prompt Engineering}
Prompt engineering focuses on designing and refining input instructions for LLMs \cite{reynolds2021prompt,shin2020autoprompt}. Recent studies have demonstrated the effectiveness of methods that  automate this search using LLMs as prompt generators \cite{zhou2023large,zhang2023automatic}. However, while LLMs are capable of producing high-quality instruction templates, at a level  comparable to  human annotators, supervision and testing remain essential to filter out poorly performing prompts.

In this section, we propose applying aLTT to automated prompt engineering  \cite{zhou2023large} to generate instructions with statistical performance guarantees. As illustrated in Figure~\ref{fig:pipeline}, given a set $\Lambda$ of candidate instructions generated by an LLM, we use aLTT to sequentially test the instructions and identify a sufficiently large number of prompts  that meet user-defined reliability requirements.

\subsubsection{Problem Definition}

\begin{figure}
\centering
\includegraphics[width=\linewidth]{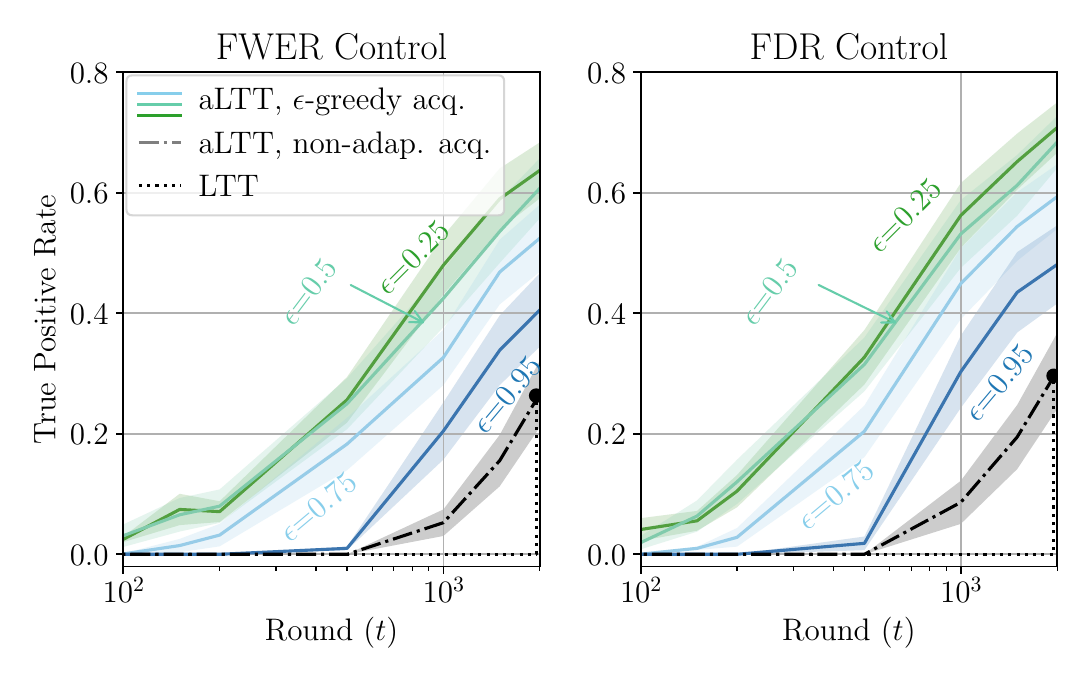}
    \caption{True positive rate of LTT and aLTT with $\epsilon$-greedy acquisition policy for $\epsilon\in\{0.25,0.5,0.75,0.95\}$ and non-adaptive acquisition. On the left panel, the prediction sets satisfy FWER control, while on the right panel they meet FDR requirements. In both cases, the tolerance level is $\delta=0.1$. Results are averaged over the tasks in \cite{honovich2022instruction} that yield non-empty reliable prompts set.}
    \label{fig:llm_1}
 \end{figure}

Consider a target LLM $f(\cdot)$ that, when fed the concatenation $[\lambda, X]$ of an instruction $\lambda$ and data $X$,  produces an output text $f([\lambda, X])$. For example, in the task of movie recommendation illustrated in Figure \ref{fig:pipeline}, 
 each prompt $\lambda$ corresponds to an instruction  to generate movie titles  similar to those provided in the input list $X$; while the output $f([\lambda, X])$ represents a list of recommended movies. Following \cite{zhou2023large,zhang2023automatic}, the candidate set $\Lambda$ of instructions, corresponding to the hyperparameters to be tested, is  generated using a separate LLM. In particular, we use the Llama3 8B Instruct LLM \cite{dubey2024llama} as the target model $f(\cdot)$, while the initial instruction set $\Lambda$ is generated through the Llama3.3 70B Instruct model \cite{llamma} using the forward generation mode detailed in \cite{zhou2023large}.

For each instruction $\lambda$, the ability of the prompted model $f([\lambda, X])$ to generate high-quality outputs for a test datum $Z = (X, Y)$ is measured by a task-dependent loss function $R(\lambda,Z)=\ell(f([\lambda, X]),Y)\in[0,1]$. Focusing on tasks from the instruction induction data set \cite{honovich2022instruction}, we specifically adopt the 0-1 loss  
tailored to the given  task \cite{zhou2023large}.  The goal  is to determine a subset $\Lambda^{\rm rel}\subseteq \Lambda$ that complies with the condition (\ref{eq:rel_set}) with a target execution error $\alpha=0.2$, and  FWER and FDR requirements given by the tolerance parameter   $\delta=0.1$.

\subsubsection{Results}
\label{sec:aep_res}

For the outlined prompt engineering problem, we evaluate aLTT using an $\epsilon$-greedy acquisition strategy with $\epsilon \in \{0.25, 0.5, 0.75, 0.95\}$, as well as a non-adaptive acquisition strategy whereby the instruction to be tested is selected randomly and independently from prior testing rounds. All schemes employ an online Newton step betting strategy \cite{waudby2024estimating} for the e-value (\ref{eq:cap_wealth}).

\begin{figure}
\includegraphics[width=\linewidth]{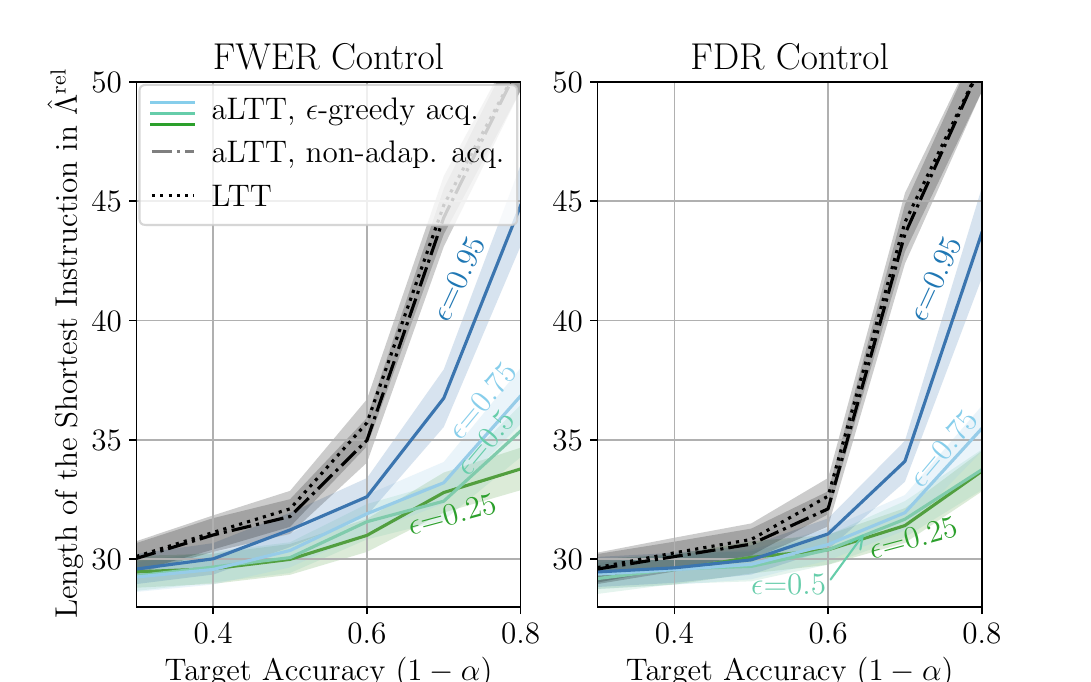}
    \caption{Length of the shortest instruction in the predicted set of reliable hyperparameter $\hat{\Lambda}^{\rm rel}$ returned by LTT and aLTT. Instructions are tested under different accuracy requirements and with a fixed testing budget $T=2000$.}
    \label{fig:llm_2}
 \end{figure}

In Figure~\ref{fig:llm_1}, we report the TPR of  LTT and aLTT as a function of the number of rounds $t$. Results are averaged over tasks that yield a final non-empty set $\hat{\Lambda}^{\rm rel}$. Plots for individual tasks are included in Appendix~\ref{app:reliable_prompt}.  Under both FDR and FWER control, aLTT reduces the number of testing rounds, or equivalently the number of LLM executions, required to achieve a given TPR. In particular,  LTT produces a non-empty set only at the end of the testing process, attaining a final TPR less than half of the TPR achieved by aLTT with  $\epsilon$-greedy acquisition strategy and $\epsilon = 0.25$. Moreover, aLTT with $\epsilon = 0.25$ identifies $50\%$ of the reliable instructions within the first $1000$ testing rounds, while the non-adaptive testing procedure identifies less than $10\%$ of the reliable instructions with the same number of rounds. 

The improved efficiency of aLTT translates into superior performance in downstream tasks. To illustrate this point, we follow the described hyperparameter selection process with  a post-selection phase that identifies a single hyperparameter $\hat{\lambda}$ from the estimated set  $\hat{\Lambda}^{\rm rel}$  of reliable instructions (see Figure 1). Specifically, we adopt  the shortest instruction $\lambda \in \hat{\Lambda}^{\rm rel}$. This post-selection criterion is motivated by the computational benefits of processing shorter prompts. 

In Figure~\ref{fig:llm_2}, we report the length of the selected instruction $\hat{\lambda}$ as a function of the target execution accuracy $1 - \alpha$. Results are averaged over tasks with a final non-empty set $\hat{\Lambda}^{\rm rel}$, while plots for individual tasks are included in Appendix~\ref{app:reliable_prompt}.  Stricter accuracy requirements are seen to reduce the number of discovered reliable instructions, leading to an increase in the minimal instruction length within $\hat{\Lambda}^{\rm rel}$. However, across all reliability levels $\alpha$, aLTT consistently delivers the shortest instructions, outperforming alternative schemes. This advantage is to be attributed to the data-adaptive acquisition strategy implemented by aLTT, which enables the discovery of a larger number of reliable instructions.}

\section{Conclusion}

We introduced aLTT, a novel framework for hyperparameter selection that implements data-dependent sequential testing via early termination. Unlike the existing LTT, which builds on p-value multiple hypothesis testing (MHT), aLTT is based on sequential MHT via e-processes \cite{xu2021unified}. In practical scenarios, including the problem of prompt engineering, this results in more efficient and flexible calibration procedures that maintain statistical validity and the same discovery power as LTT by using only a fraction of testing rounds.

Potential extensions of the aLTT framework include the study of scenarios characterized by distribution shift and simulation-aided calibration.

\section{Acknowledgments}
This work was supported by the European Union’s Horizon Europe project CENTRIC (101096379). The work of Osvaldo Simeone was also supported by the Open Fellowships of the EPSRC (EP/W024101/1) and by the EPSRC project  (EP/X011852/1).
Thanks also to the Advanced Research and Invention Agency (ARIA) for supporting broader foundational work in this space.
\bibliography{biblio}
\bibliographystyle{icml2025}

  
\newpage
\appendix
\onecolumn

\section{FWER and FDR-Controlling Procedures}
In this section we review popular FWER and FDR-controlling strategies applied to the problem of hyperparameter selection.
\vspace{-1em}
\subsection{FWER Control}
\subsubsection{Bonferroni Correction}
For a set of p-values $\mathcal{P}=\{P_i\}^N_{i=1}$, a simple $(\alpha,\delta)$-FWER-controlling procedure is given by the Bonferroni correction $\mathcal{A}^{\rm FWER}_{\rm Bon}(\mathcal{P})$, which returns the hyperparameter set 
\begin{align}
	\hat{\Lambda}^{\rm rel}_{\rm Bon}=\left\{\lambda_i: P_i\leq \frac{\delta}{N}\right\}.
	\label{eq:bonferroni}
\end{align}
The hyperparameter set $\hat{\Lambda}^{\rm rel}_{\rm Bon}$ is guaranteed to be $\delta$-FWER-controlling \cite{1570009749360424576}.
\subsubsection{Fixed-Sequence Testing}
If the designer has access to prior knowledge $\mathcal{D}^0$ about which hyperparameters are likely to be more reliable, it is possible to increase the power of the statistical test, and thus the cardinality of the hyperparameter $\hat{\Lambda}^{\rm rel}$, by using a \emph{fixed-sequence testing} procedure $\mathcal{A}^{\rm FWER}_{\rm FS}(\mathcal{P},\mathcal{D}^0)$. In fixed sequence hypothesis testing, the hypotheses $\{\mathcal{H}_i\}_{i=1}^N$ are ordered based on the prior knowledge $\mathcal{D}^0$ from the most likely to be reliable to the least likely to be reliable. Denote the $k$-th hypothesis in the corresponding ordered sequence as $\mathcal{H}_{(k)}$, and the associated p-value as $P_{(k)}$. With fixed-sequence testing, the hypotheses are sequentially tested at a reliability level $\alpha$ until one is accepted. Accordingly, the resulting  hyperparameter set contains all hypotheses up until the first acceptance, i.e.,
\begin{align}
	\hat{\Lambda}^{\rm rel}_{\rm FS}=\left\{\lambda_{(j)}: P_{(i)}\leq \delta,\ \forall i \leq j \right\}.
	\label{eq:fixed_seq}
\end{align}
For any ordering of the p-values, the hyperparameter set $\hat{\Lambda}^{\rm rel}_{\rm FS}$ is guaranteed to be $\delta$-FWER-controlling \cite{bauer1991multiple}.

\subsection{FDR Control}
\subsubsection{Benjamini-Hochberg Procedure}
Given a set of independent p-values $\mathcal{P}$, denote the $k$-th smallest value in the set as $P_{(k)}$ and the associated hyperparameter as $\lambda_{(k)}$. For an error level $\delta$, the Benjamini-Hochberg procedure $\mathcal{A}^{\rm FDR}_{\rm BH}(\mathcal{P})$ returns the prediction set

\begin{align}
	\hat{\Lambda}^{\rm rel}_{\rm BH}=\left\{\lambda_{(i)}:P_{(i)} \leq \frac{i\delta}{N} \right\}.
	\label{eq:bh}
\end{align}

By (\ref{eq:bh}), BH applies a larger threshold to hyperparameters $\lambda\in\Lambda$ that have larger p-values and are thus less likely to be reliable. If the p-values in set $\mathcal{P}$ are independent, the Benjamini-Hochberg (BH) procedure $\mathcal{A}^{\rm FDR}_{\rm BH}(\mathcal{P})$ guarantees FDR control at a level $\delta$ \cite{benjamini1995controlling}. 

\subsubsection{Benjamini-Yekutieli Procedure}
In case of arbitrarily dependent p-values in set $\mathcal{P}$ the error level $\delta$ has to be adjusted by a multiplicative factor $(\sum^N_{n=1}1/n)^{-1}$. The resulting FDR-controlling procedure, also known as the Benjamini-Yekutieli (BY) procedure, yields the set

\begin{align}
	\hat{\Lambda}^{\rm rel}_{\rm BY}=\left\{\lambda_{(i)}:P_{(i)} \leq \frac{i\delta}{\sum^N_{n=1}N/n} \right\}.
	\label{eq:by}
\end{align}
The BY  procedure $\mathcal{A}^{\rm FDR}_{\rm BY}(\mathcal{P})$ is $\delta$-FDR-controlling \cite{benjamini2001control}.
\subsubsection{E-Benjamini-Hochberg Procedure}
Alternatively, given the set of e-values $\mathcal{E}^t=\{E_i^t\}^N_{i=1}$, denote by the $k$-th largest e-value in $\mathcal{E}$ as $E^t_{(k)}$  and the associated hyperparameter as $\lambda_{(k)}$. For an error level $\delta$, the e-Benjamini-Hochberg procedure $\mathcal{A}^{\rm FDR}_{\rm eBH}(\mathcal{E})$ outputs the hyperparameter subset
\begin{align}
	\hat{\Lambda}^{\rm rel}_{\rm eBH}=\left\{\lambda_{(i)}: E^t_{(i)}\geq \frac{N}{i\delta}\right\}.
	\label{eq:ebh}
\end{align}

Following the same basic principles underlying BH in (\ref{eq:bh}), eBH applies a smaller threshold to hyperparameters $\lambda\in\Lambda$ that have smaller e-values and are thus less likely to be reliable.
The eBH procedure returns an $\delta$-FDR-controlling set even for arbitrarily dependent e-values $E^t_{(k)}$ without the need to adjust the error level $\delta$ as in the case of the p-value-based BH procedure \cite{xu2021unified}.

\section{Additional Experiments}
\subsection{Online Policy Selection for Offline Reinforcement Learning}
\subsubsection{Effect of the Betting Strategy}
\begin{figure}
	\centering
	\includegraphics[width=0.5\linewidth]{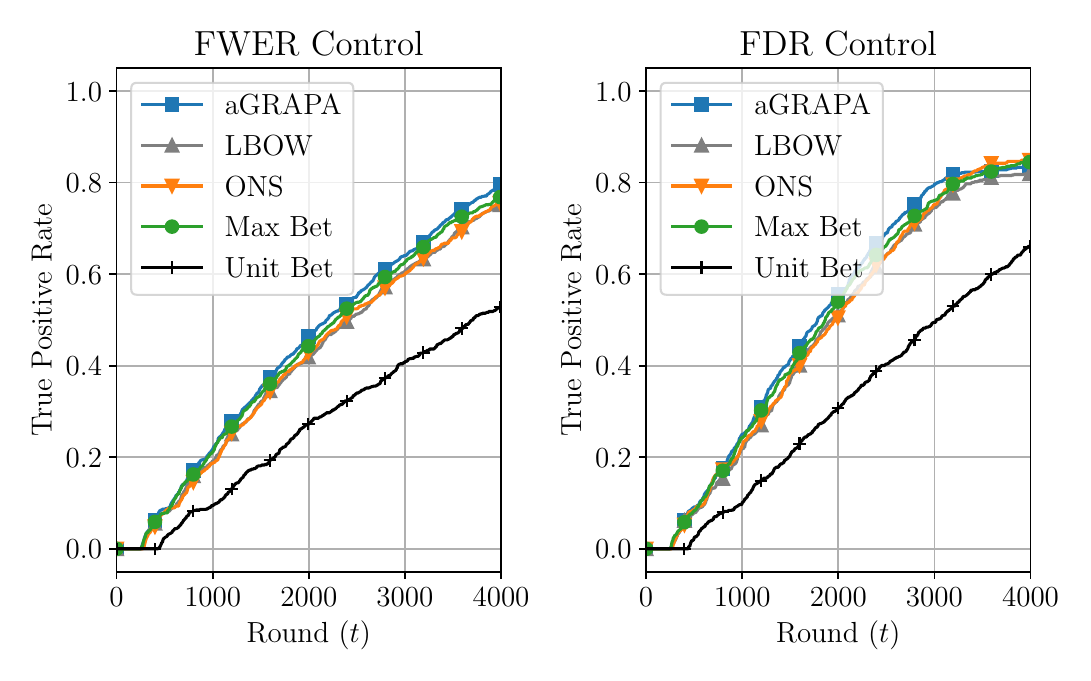}  
	\caption{True positive rate of aLTT with $\epsilon$-greedy acquisition policy with $\epsilon=0.25$ under different betting strategies. On the left panel, the prediction sets satisfy FWER control while on the right FDR control. In both cases, the tolerance level is $\delta=0.1$.}
	\label{fig:betting}
\end{figure}
In Section 5.1, we studied the TPR of aLTT under the \emph{approximate growth rate adaptive to the particular alternative} (aGRAPA) betting strategy. In the following, we compare its performance against non-adaptive betting strategies, specifically the \emph{maximum bet} (MaxBet) strategy, which sets the bet $\mu^{t}_i$ to the maximum allowed value of $1/\alpha$, and the \emph{unitary bet} (UnitBet) strategy, where $\mu^{t}_i = 1$. Additionally, we evaluate two alternative adaptive online betting strategies based on approximate wealth maximization \cite{waudby2024estimating}: \emph{lower-bound on the wealth} (LBOW) and \emph{online Newton step} (ONS).

Under the same online policy selection for offline reinforcement learning as set up in Section 5.1, in Figure \ref{fig:betting} we compare the TPR of aLTT with an $\epsilon$-greedy acquisition policy. The left panel targets FWER control, while the right focuses on FDR control. All adaptive betting strategies exhibit similar performance, with aGRAPA showing a slight advantage. Interestingly, in this scenario, the non-adaptive MaxBet strategy performs surprisingly well, while UnitBet achieves the lowest TPR.

\subsubsection{Quantile Risk Control}
\begin{figure}
	\centering
\includegraphics[width=0.5\linewidth]{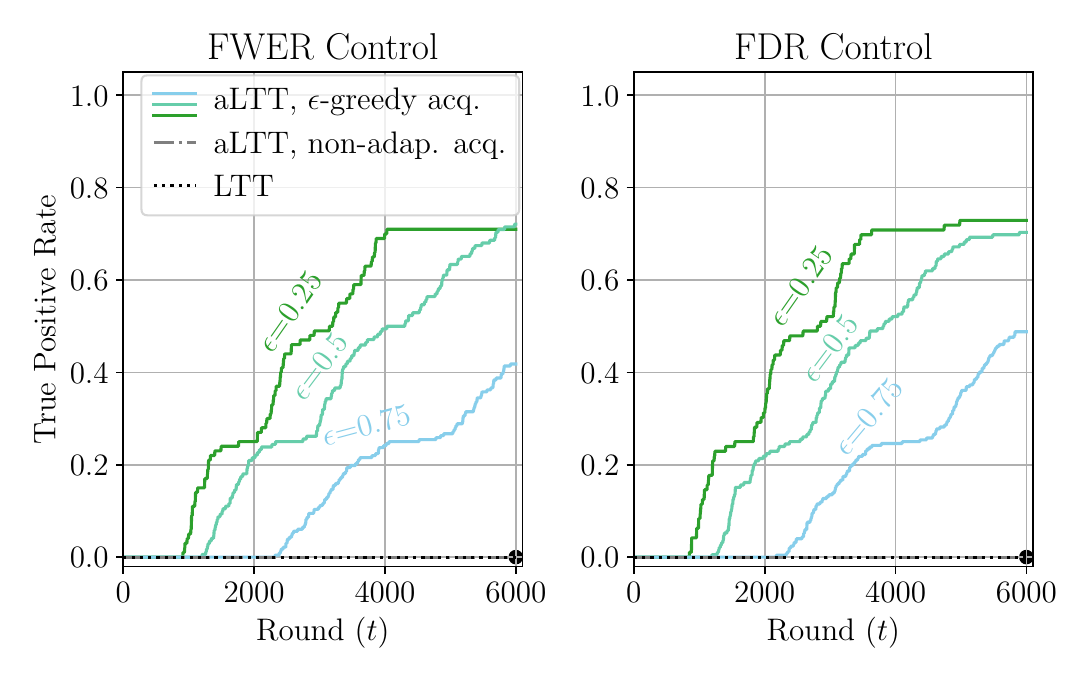} 
     \vspace{-1em}
	\caption{True positive rate of LTT and aLTT with $\epsilon$-greedy acquisition policy for $\epsilon\in\{0.25,0.5,0.75,0.95\}$ and non-adaptive acquisition. On the left panel, the prediction sets satisfy FWER control while on the right FDR control. In both cases, a policy is considered reliable if the quantile risk at level $q=0.1$ is lower than $\alpha=0.57$, and the tolerance level is $\delta=0.1$.}    \label{fig:rl_quantile}	  \vspace{-1em}
\end{figure}
{\color{black}
In this experiment, we consider the same set-up of Section 5.1 with the caveat that, instead of identifying policies with a large enough average reward, we focus on policies with a small enough quantile risk --- a risk definition that explicitly captures safety \cite{dabney2018distributional,farzaneh2024quantile}. The quantile risk of a hyperparameter $\lambda$ at a level $q$ is defined as
\begin{align}
    R_q(\lambda)=\inf_r\{r:\Pr[R(\lambda, Z)\leq r]>1-q\}.
\end{align}

Controlling the quantile risk ensures guarantees on the average fraction of times the policy risk exceeds the level $R_q(\lambda)$. For the RL setting, we deal with rewards, so the inequalities in the definition are reversed.

We aim to identify policies with a quantile risk at the level $q = 0.1$ that is smaller than $\alpha = 0.57$. In Figure \ref{fig:rl_quantile}, we report the true positive rate of aLTT using the $\epsilon$-greedy acquisition policy and LTT. aLTT significantly outperforms LTT, identifying up to 75\% of the policies that satisfy the given reliability requirement within the maximum testing round, $T$. In contrast, LTT is unable to identify any policy due to its non-adaptive testing strategy.}

\subsection{Reliable Hyperparameter Selection for Wireless Resource Allocation}

In wireless communication systems, resource allocation is an essential functionality that regulates access to the spectrum for users and services \cite{stanczak2009fundamentals}. The performance of resource allocation policies is evaluated by using key performance indicators (KPIs) such as throughput, delay, and energy efficiency. Despite the randomness inherent in the network conditions, some services require strict reliability guarantees in terms of KPIs. For instance, gaming applications must meet latency constraints \cite{elbamby2019wireless}, streaming connections are subject to throughput requirements \cite{li2012maximizing}, and battery-powered transmitters have strict energy-efficiency constraints \cite{mahapatra2015energy}.

\subsubsection{Problem Definition}

\begin{figure}
	\centering
	\includegraphics[width=0.45\linewidth]{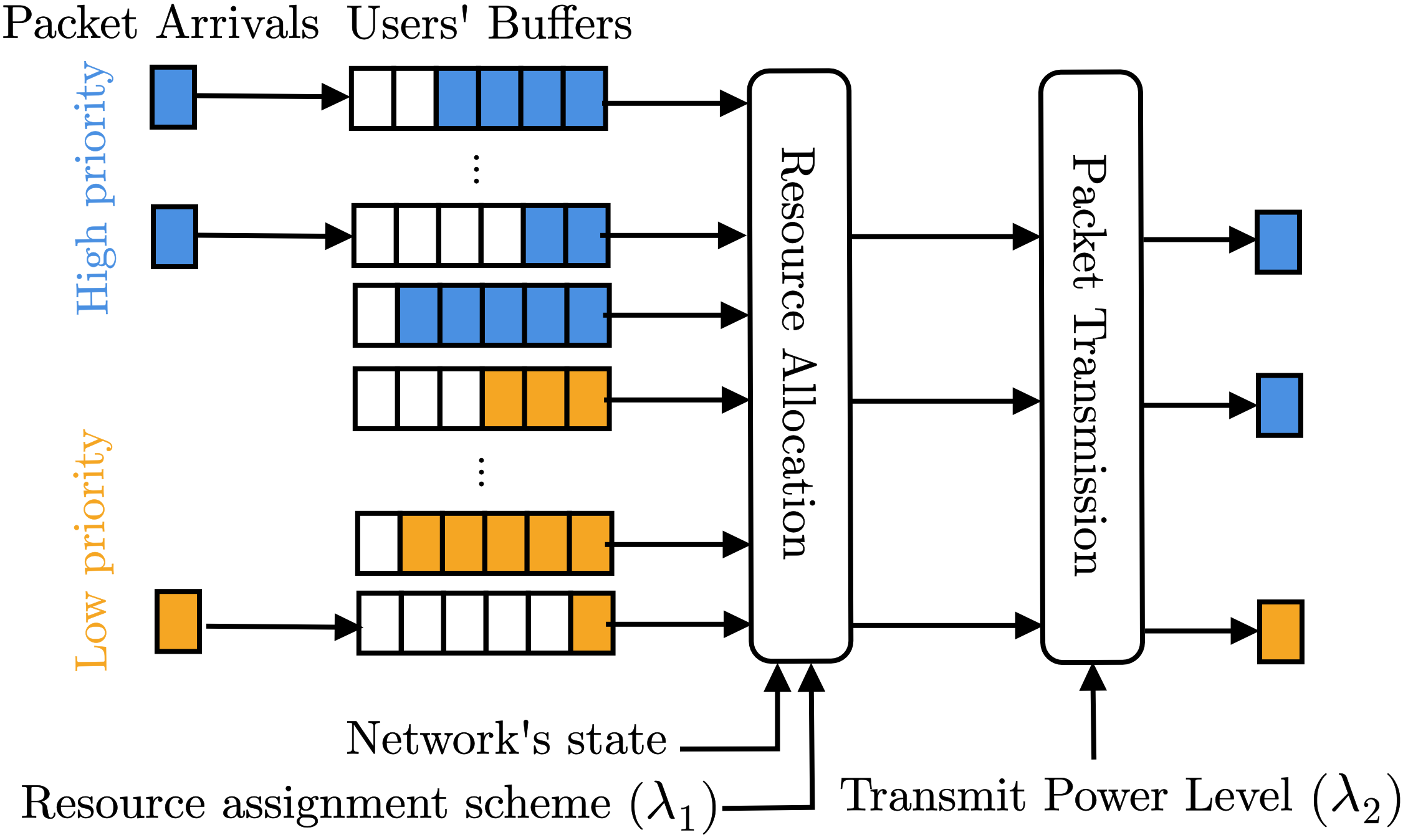}  
	\caption{System diagram of the considered resource allocation problem. At each transmission frame  $k$, new packets are randomly generated and stored in the corresponding users' buffers. A resource allocation scheme, determined by hyperparameter  $\lambda_1$,  assigns resources to the users, deciding whose users to schedule at round $k$. Packets from the scheduled users’ queues are then transmitted over a wireless channel using a transmit power dictated by the hyperparameter  $\lambda_2$.}
	\label{fig:system}
  \vspace{-1em}
\end{figure}

As illustrated in Figure \ref{fig:system}, we consider a downlink resource allocation problem in which, at every transmission frame $k\geq 1$, a base station serves users by communicating bits from the corresponding queues. This is done by assigning physical resource blocks (PRBs) to users based on a descriptor $s_k$ of the network conditions that includes the state of users’ transmission buffers, users' priorities, and channel conditions. Following the simulation software Nokia Wireless Suite \cite{wirelesssuite}, we consider three different types of PRBs assignment policies: a proportional fair scheme \cite{baruah1993proportionate}, a knapsack allocation policy \cite{ferdosian2016greedy}, and a learning-based scheme based on a Covariance Matrix Adaptation Evolution Strategy (CMA-ES) \cite{de2020radio}.  Overall, a resource allocation scheme is specified by the hyperparameters $\lambda=\{\lambda_1,\lambda_2\}$, with $\lambda_1\in\{\text{proportional fair, knapsack, CMA-ES}\}$ identifying the scheduling policy, and $\lambda_2$ determining the transmit power level. We let the hyperparameter $\lambda_2$ vary from 0 dB to 24 dB in 1 dB increments. This way, the set $\Lambda$ contains 75 candidate hyperparameters.

An instance of the resource allocation problem is described by the random quantity  $Z$, which encompasses channel conditions and packet generation.  Furthermore, the performance of the resource allocation hyperparameter $\lambda$ is measured by the following KPIs: \emph{(i)} \textit{average transmission delay} of a packet $T_{\rm tx}(\lambda,Z)$; \emph{(ii)} overall \emph{energy efficiency}, which is defined as the ratio between the number of transmitted bits and the overall transmitted energy within the episode, $\rho(\lambda,Z)=\sum_{k=1}^K B^k/\sum_{k=1}^K E^k$
where $B^k$ and $E^k$ are the number of transmitted bits and transmit energy at slot $k$; \emph{(iii)}  \emph{average queue length occupancy} $Q(\lambda,Z)$, which is normalized by the maximum queue size $Q_{\rm max}$; \emph{(iv)} average \emph{energy-delay product},
${\rm EDP}(\lambda,Z)=T_{\rm tx}(\lambda,Z)\sum_{k=1}^K E^k/K$,
quantifying the overall performance in terms of energy and delay \cite{laros2013energy}.

\subsubsection{Results}

\begin{figure}
	\centering
	\includegraphics[width=0.5\linewidth]{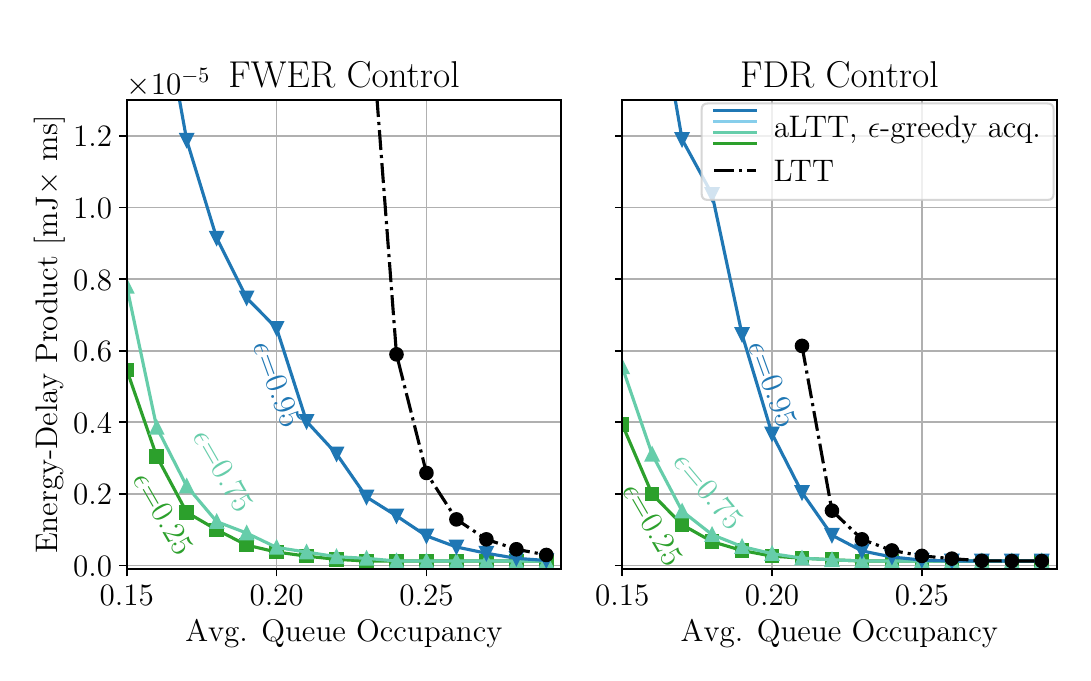} 
     \vspace{-1em}
	\caption{Average energy-delay product of the resource allocation policy returned by LTT and aLTT with $\epsilon$-greedy acquisition policy with $\epsilon\in\{0.25,0.5,0.95\}$ under different queue occupancy efficiency requirements. The final hyperparameter is obtained by selecting the hyperparameter in the estimated set of reliable policies $\hat{\Lambda}^{\rm rel}$ that is associated with the lowest empirical energy-delay product. On the left panel, the prediction sets satisfy FWER control, while on the right FDR control, with the tolerance level $\delta=0.1$. }
	\label{fig:two_classes}
  \vspace{-1em}
\end{figure}

We study a setting with differentiated service requirements. In particular, we enforce a constraint on the average queue occupancy and energy efficiency for high-priority users, while making a best-effort attempt at minimizing the energy-delay product for low-priority users.  Writing as $Q^{\rm HI}(\lambda)=\mathbb{E}_{P_Z}[Q^{\rm HI}(\lambda,Z)]$ the average queue occupancy of high-priority users, and as $\rho^{\rm HI}=\mathbb{E}_{P_Z}[\rho^{\rm HI}(\lambda,Z)]$ the average energy efficiency of high-priority users, and as $(\alpha_1,\alpha_2)$ the corresponding requirements. A hyperparameter is reliable if it meets both the queue occupancy and energy efficiency requirements for high-priority users, i.e., $\Lambda^{\rm rel}=\{\lambda\in\Lambda: Q^{\rm HI}(\lambda)\leq \alpha_1 \cap \rho^{\rm HI}(\lambda)\geq \alpha_2\}$. An e-process for the hypothesis $\mathcal{H}_i:Q^{\rm HI}(\lambda_i)> \alpha_1 \cap \rho^{\rm HI}(\lambda_i)< \alpha_2$ can be obtained as the minimum between an e-process for hypothesis $\mathcal{H}^{1}_i:Q^{\rm HI}(\lambda_i)> \alpha_1$, and an e-process for the hypothesis $\mathcal{H}^{2}_i:\rho^{\rm HI}(\lambda_i)< \alpha_2$. {\color{black}Note that any e-merging function can be applied to ensure validity over the joint hypothesis (see, e.g., \citep{vovk2021values, ramdas2024hypothesis}), we choose minimum for simplicity.}

In Figure \ref{fig:two_classes}, we compare the performance of LTT and aLTT for a calibration horizon $T=4000$ by choosing in the estimated set of reliable hyperparameters $\hat{\Lambda}^{\rm rel}$ the hyperparameter $\hat{\lambda}$ that minimizes the empirical energy-delay product of low-priority users based on all the data $\mathcal{D}^T$ collected across the testing steps. We vary the queue occupancy reliability level $\alpha_1$ while fixing the energy-efficiency requirement $\alpha_2=0.01$ Mbit/Joule. As the queue occupancy target $\alpha_1$ increases, the requirement becomes less stringent, and the energy-delay product of the low-priority users under the returned policy decreases. In both cases—FWER control on the left and FDR control on the right—the performance of aLTT is significantly superior to that of LTT. What is more, in this example, the looser guarantees provided by FDR do not entail any reliability loss, since the FWER of the hyperparameter  $\hat{\lambda}$ returned by FDR-controlling procedures was found to be significantly below $\delta=0.1$.

\subsubsection{Minimizing Average Delay under an Energy Efficiency Requirement}

Now we consider the wireless resource allocation problem under a reliability requirement on the average energy efficiency. Specifically, we consider the population risk $R(\lambda)=\mathbb{E}_{P_Z}[\rho(\lambda,Z)]$, so that the set of reliable hyperparameters is given by $\Lambda^{\rm rel}=\{\lambda\in\Lambda: R(\lambda)\geq \alpha\}$ for the given target energy efficiency $\alpha$. After $T=1000$ testing steps, with each step consisting of a resource allocation episode of $K=2500$ frames, the hyperparameter selection scheme selects from the estimated subset of reliable hyperparameters $\hat{\Lambda}^{\rm rel}$, the hyperparameter $\hat{\lambda}$ that minimizes the empirical communication delay $\hat{T}_{\rm tx}(\lambda,\mathcal{D}^T)=1/|\mathcal{D}^T|\sum_{Z\in \mathcal{D}^T}\hat{T}_{\rm tx}(\lambda,Z)$ based on all the data $\mathcal{D}^T$ collected across the testing steps.

\begin{figure}
	\centering
	\includegraphics[width=0.5\linewidth]{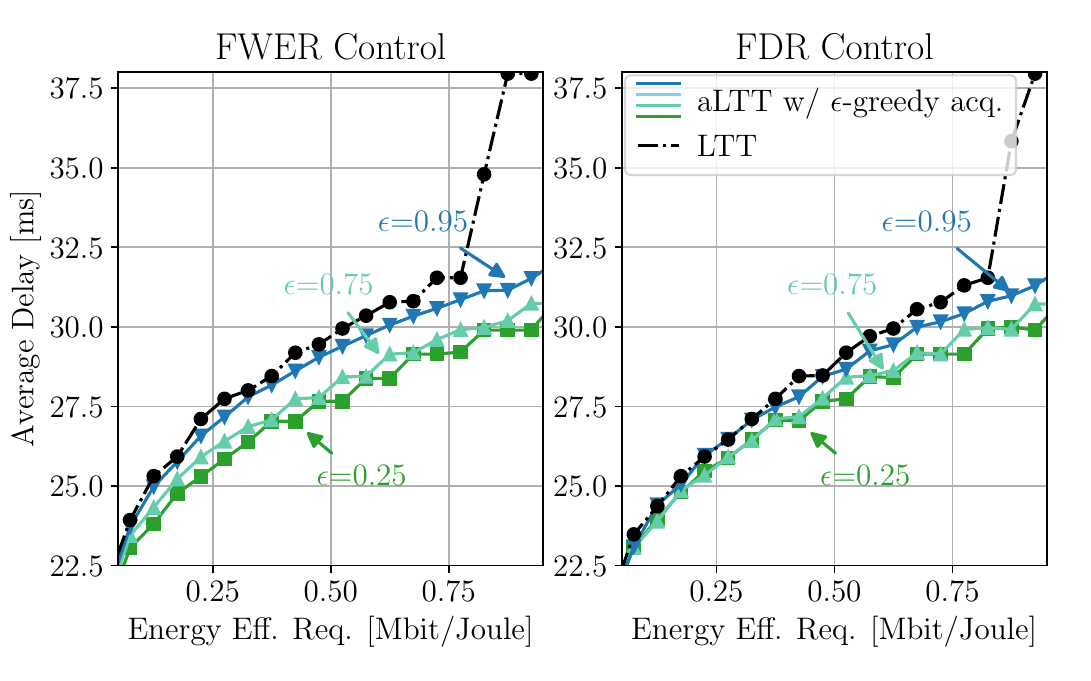}  
	\caption{Average delay of the resource allocation policy returned by LTT and aLTT with $\epsilon$-greedy acquisition policy with $\epsilon\in\{0.25,0.5,0.95\}$ under different energy efficiency requirements. The final policy is obtained by selecting the policy in the estimated set of reliable policies $\hat{\Lambda}^{\rm rel}$ that is associated to the lowest empirical delay. On the left panel, the prediction sets satisfy FDR control, while on the right FWER control, with the tolerance level $\delta=0.1$. }
	\label{fig:all_eevsdelay}
\end{figure}

In Figure \ref{fig:all_eevsdelay} we report the average delay $T_{\rm tx}(\hat{\lambda})=\mathbb{E}_{P_Z}[T_{\rm tx}(\hat{\lambda},Z)]$, estimated on hold-out data, of the selected policy $\hat{\lambda}$ as a function of the reliability level $\alpha$ for LTT and aLTT with an $\epsilon$-greedy acquisition function, when setting the target error level $\delta=0.1$. On the left panel, we consider FWER control, while the right presents the performance with FDR control. 

As the reliability requirement $\alpha$ increases, becoming more stringent, the average delay of the selected hyperparameter $\hat{\lambda}$ increases. However, thanks to adaptive testing, aLTT returns a larger prediction set $\hat{\Lambda}^{\rm rel}$ providing hyperparameters $\hat{\lambda}$ with lower delays as compared to LTT. Furthermore, due to the less stringent FDR requirements, FDR-controlling testing procedures generally return policies with average delays smaller than FWER-controlling. 

{\color{black}
\subsection{Reliable Prompt Selection for Natural Language Processing} \label{app:reliable_prompt}

In Figures \ref{fig:llm_1} and \ref{fig:llm_2}, we report the performance averaged over ten tasks from the instruction induction dataset \cite{honovich2022instruction}. Complementing the results in the main text, here we provide a breakdown of the methods’ performance across the ten tasks. In Figure \ref{fig:llm_1_app}, we show the TPR of the schemes considered in Section \ref{sec:aep_res} under FDR and FWER control with tolerance $\delta=0.1$. Across all tasks, aLTT consistently outperforms LTT, which returns empty sets for many tasks. 

In Figure \ref{fig:llm_2_app}, we plot the length of the shortest instruction in the predicted set $\hat{\Lambda}^{\text{rel}}$ returned by the different hyperparameter selection schemes as a function of the execution accuracy $1 - \alpha$. If a prediction scheme returns an empty predicted hyperparameter set, we set the instruction length to the length of the longest instruction in the candidate set $\Lambda$. For all tasks, aLTT can identify shorter instructions compared to LTT under the same testing budget.

\begin{figure}
	\centering
        \includegraphics[width=0.9\textwidth]{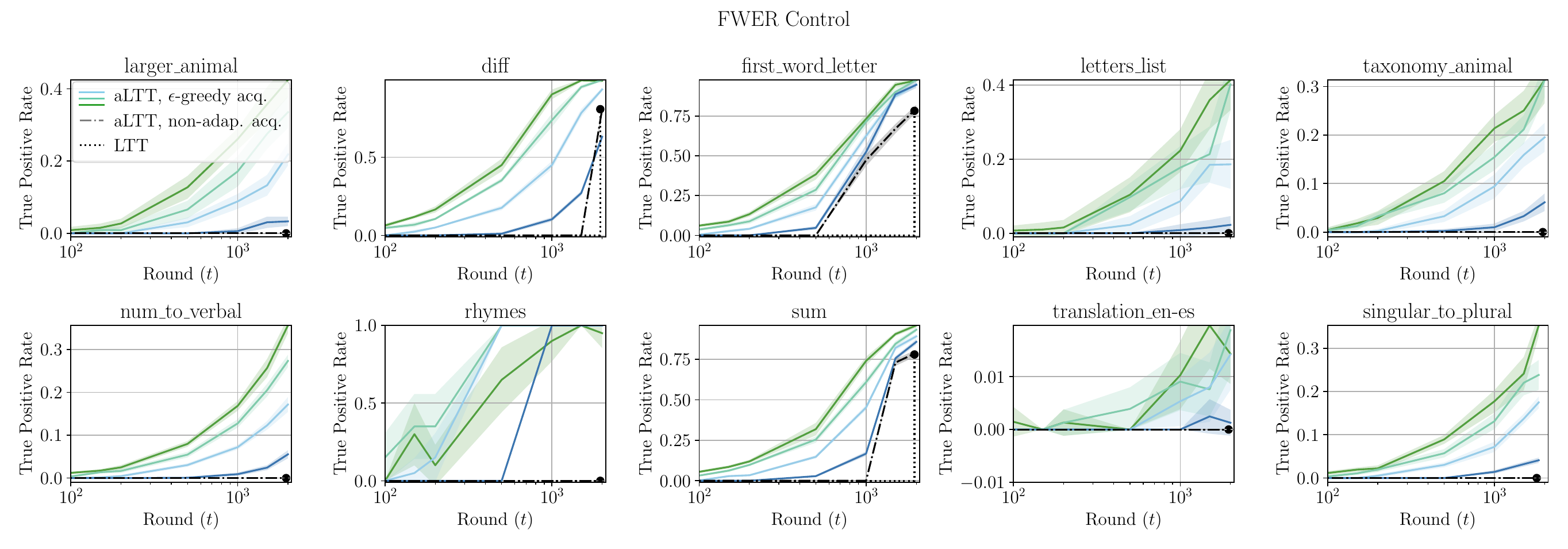}
    \includegraphics[width=0.9\textwidth]{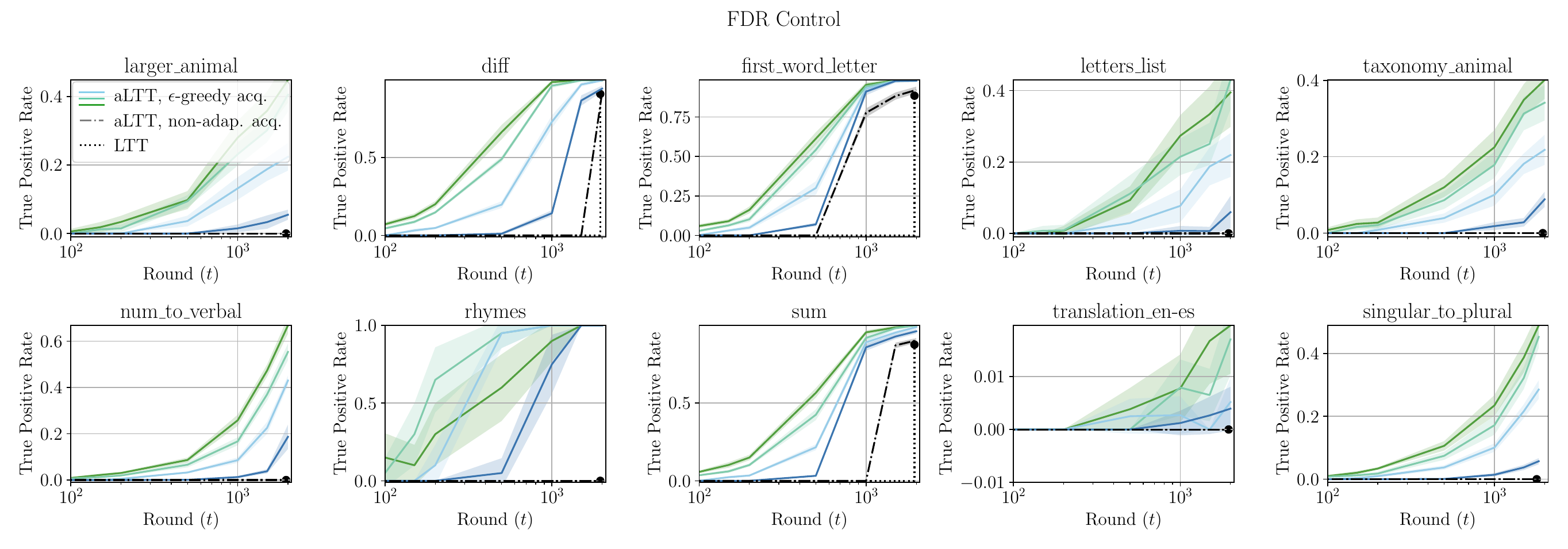}
     \vspace{-1em}
	\caption{Breakdown of the average performance of methods presented in Figure \ref{fig:llm_1} across the 10 tasks in the instruction induction dataset \cite{honovich2022instruction} for which a non-empty hyperparameter sets $\hat{\Lambda}^{\text{rel}}$ is returned. Across all tasks, aLTT delivers the highest true positive rate.}    \label{fig:llm_1_app}	  \vspace{-1em}
\end{figure}

\begin{figure}
	\centering
    \includegraphics[width=0.9\textwidth]{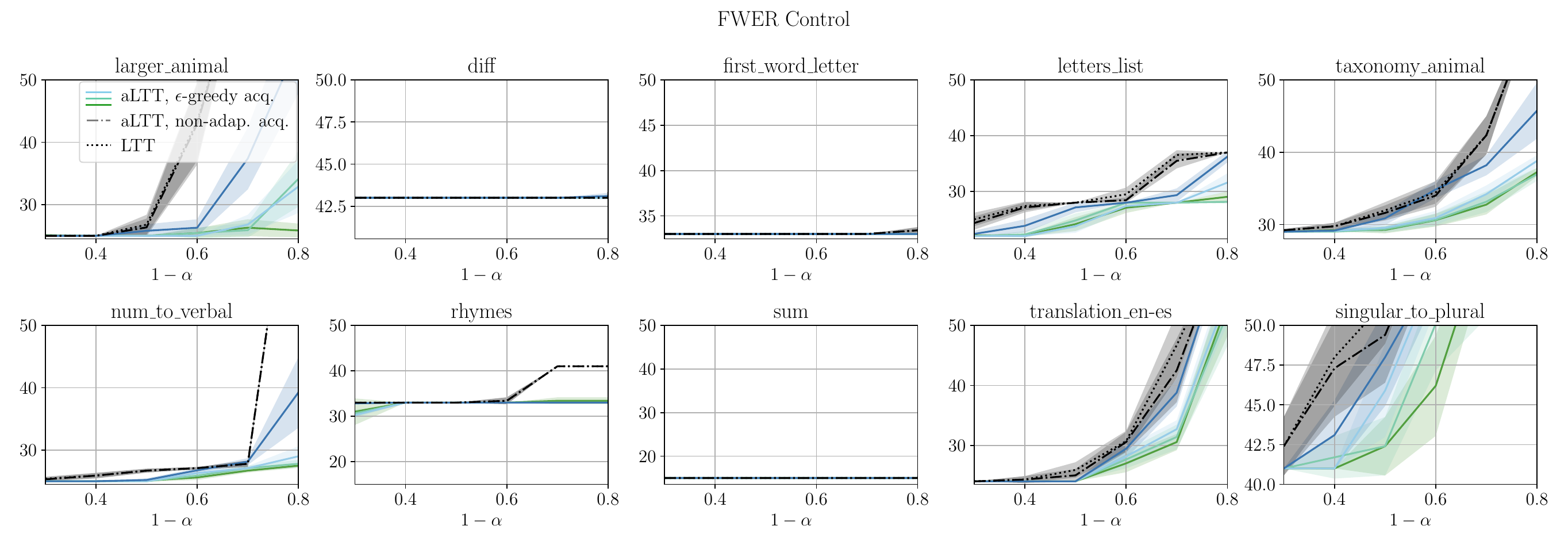}
    \includegraphics[width=0.9\textwidth]{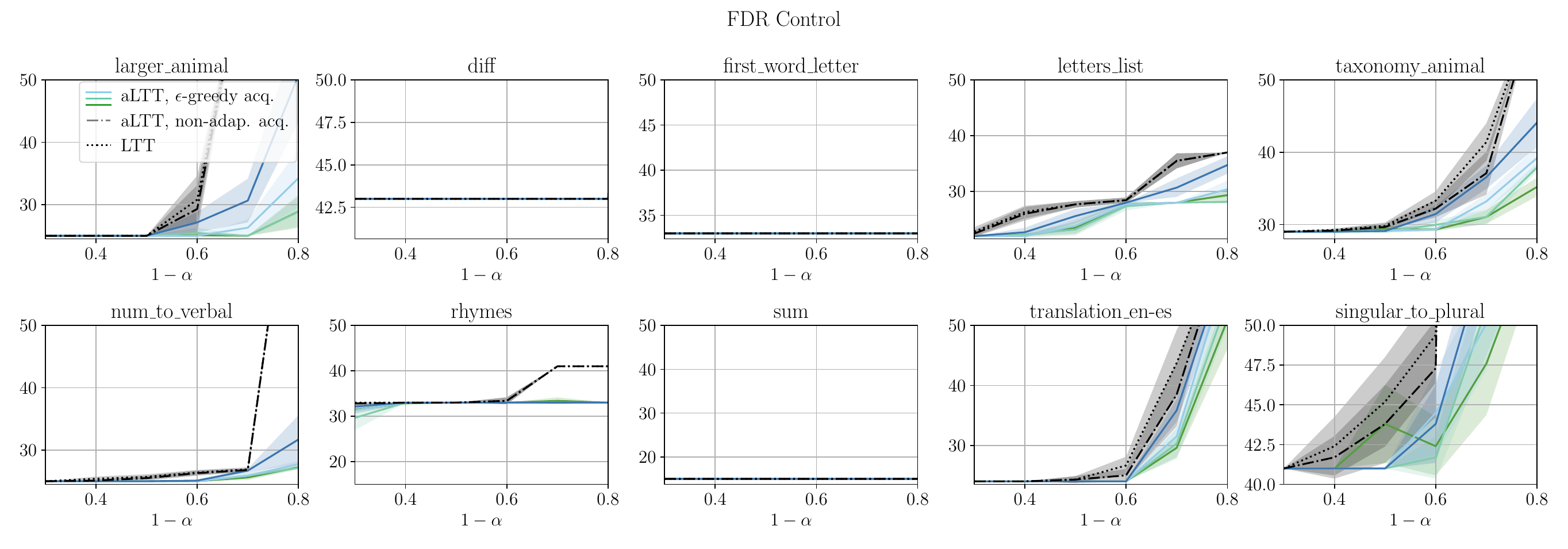}
     \vspace{-1em}
	\caption{ Length of the shortest instruction in $\hat{\Lambda}^{\text{rel}}$ as a function of target execution accuracy $1-\alpha$ for each task in the instruction induction dataset \cite{honovich2022instruction} for which the methods presented in Figure \ref{fig:llm_2} return non-empty hyperparameter sets $\hat{\Lambda}^{\text{rel}}$.}  \label{fig:llm_2_app}	  \vspace{-1em}
\end{figure}

}

\end{document}